\documentclass[conference]{IEEEtran}
\IEEEoverridecommandlockouts
\usepackage{cite}
\usepackage{amsmath,amssymb,amsfonts}
\usepackage{algorithmic}
\usepackage{graphicx}
\usepackage{textcomp}
\usepackage{xcolor}
\usepackage{hyperref,booktabs,multirow,array,color,colortbl, float, cleveref }
\hypersetup{hidelinks} 

\def\BibTeX{{\rm B\kern-.05em{\sc i\kern-.025em b}\kern-.08em
    T\kern-.1667em\lower.7ex\hbox{E}\kern-.125emX}}
\begin{document}

\title{Multi-Scale Heterogeneity-Aware Hypergraph Representation for Histopathology Whole Slide Images}

\author{
    \IEEEauthorblockN{Minghao Han$^1$, Xukun Zhang$^1$, Dingkang Yang$^1$, Tao Liu$^1$, Haopeng Kuang$^1$, Jinghui Feng$^1$, Lihua Zhang$^{1,2,3,4,*}$}
    \thanks{*Corresponding author. \newline \indent This project was funded by the National Natural Science Foundation of China 82090052.}
    \IEEEauthorblockA{$^1$ Academy for Engineering and Technology, Fudan University, Shanghai, China}
    \IEEEauthorblockA{$^2$ Changchun Boli Technologies Co., Ltd., Changchun, China}
    \IEEEauthorblockA{$^3$ Engineering Research Center of AI and Robotics, Ministry of Education, Shanghai, China}
    \IEEEauthorblockA{$^4$ Intelligent Science and Engineering Joint Key Laboratory of Jilin Province, China}
    \IEEEauthorblockA{\texttt{mhhan22@m.fudan.edu.cn, lihuazhang@fudan.edu.cn}}
}
\maketitle

\begin{abstract}
Survival prediction is a complex ordinal regression task that aims to predict the survival coefficient ranking among a cohort of patients, typically achieved by analyzing patients' whole slide images. Existing deep learning approaches mainly adopt multiple instance learning or graph neural networks under weak supervision. Most of them are unable to uncover the diverse interactions between different types of biological entities(\textit{e.g.}, cell cluster and tissue block)  across multiple scales, while such interactions are crucial for patient survival prediction. In light of this, we propose a novel multi-scale heterogeneity-aware hypergraph representation framework. Specifically, our framework first constructs a multi-scale heterogeneity-aware hypergraph and assigns each node with its biological entity type. It then mines diverse interactions between nodes on the graph structure to obtain a global representation. Experimental results demonstrate that our method outperforms state-of-the-art approaches on three benchmark datasets. Code is publicly available at \href{https://github.com/Hanminghao/H2GT}{https://github.com/Hanminghao/H2GT}.
\end{abstract}


\section{Introduction}
In light of the rapid advancement in deep learning technologies~\cite{yang2024robust,yang2023context,yang2022emotion,yang2022disentangled,yang2022learning, zhang2023anatomical}, the importance of artificial intelligence in the medical field is steadily gaining recognition. This prominence is particularly evident in the realm of survival prediction, a crucial practice in clinical procedures, which largely anticipates patient survival coefficients based on specific indicators or factors. A higher survival coefficient indicates a longer post-treatment survival period for the patient. In clinical practice, survival prediction is commonly conducted by analyzing patients' Whole Slide Images (WSIs), which serve as the ``gold standard" for cancer diagnosis, staging, and survival prediction~\cite{ludwig2005biomarkers,glaser2017light,qu2022towards,lu2024visual}. By developing survival prediction models based on WSIs, valuable references for clinical decision-making can be obtained, consequently guiding future therapeutic research.
\begin{figure}[tb]
    \centering
    \vspace{0cm}
    \includegraphics[width=1\columnwidth]{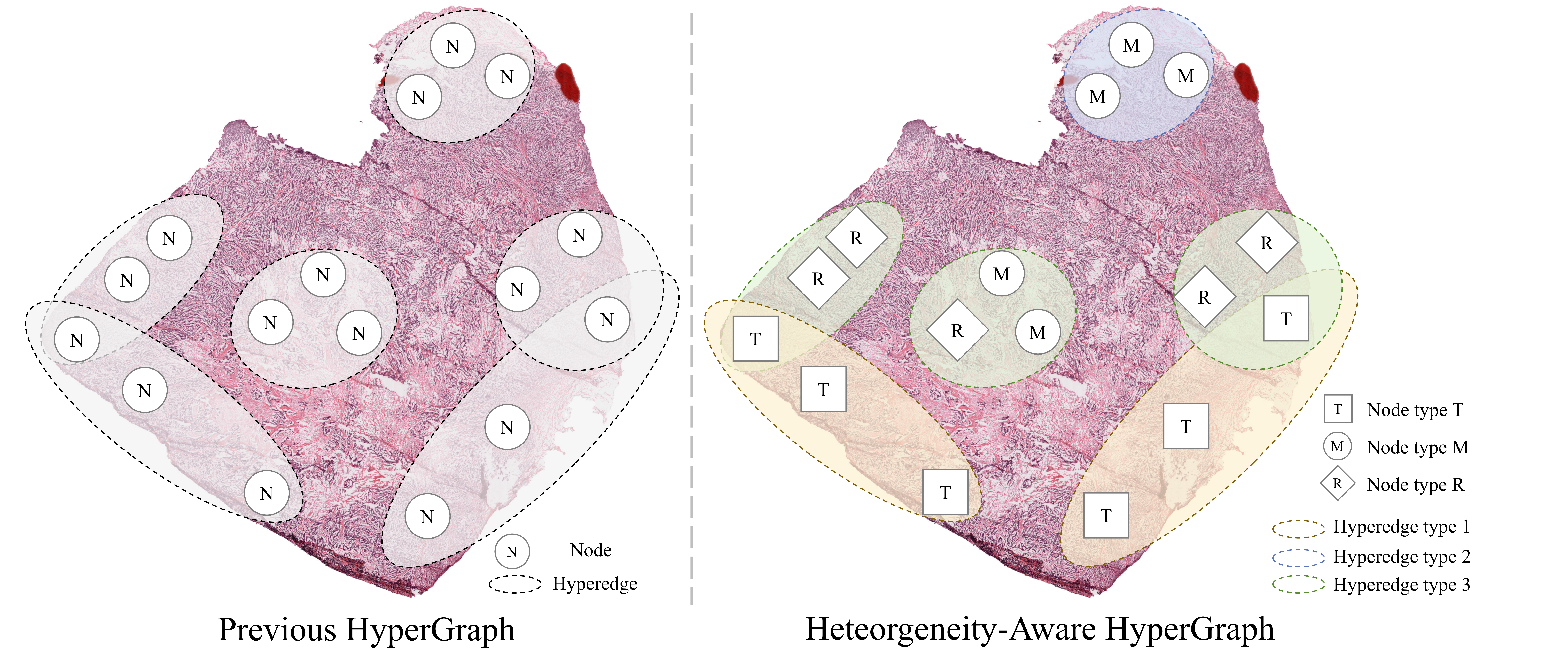}
    \caption{\textbf{Left}: Previous Hypergraphs only contain one type of node and hyperedge.\quad  \textbf{Right}: The Heteorgeneity-Aware HyperGraphs contain types of nodes and hyperedges.}
    \label{fig1}
\end{figure}

\begin{figure*}[!htp]
    \centering
    \includegraphics[width=2.0\columnwidth]{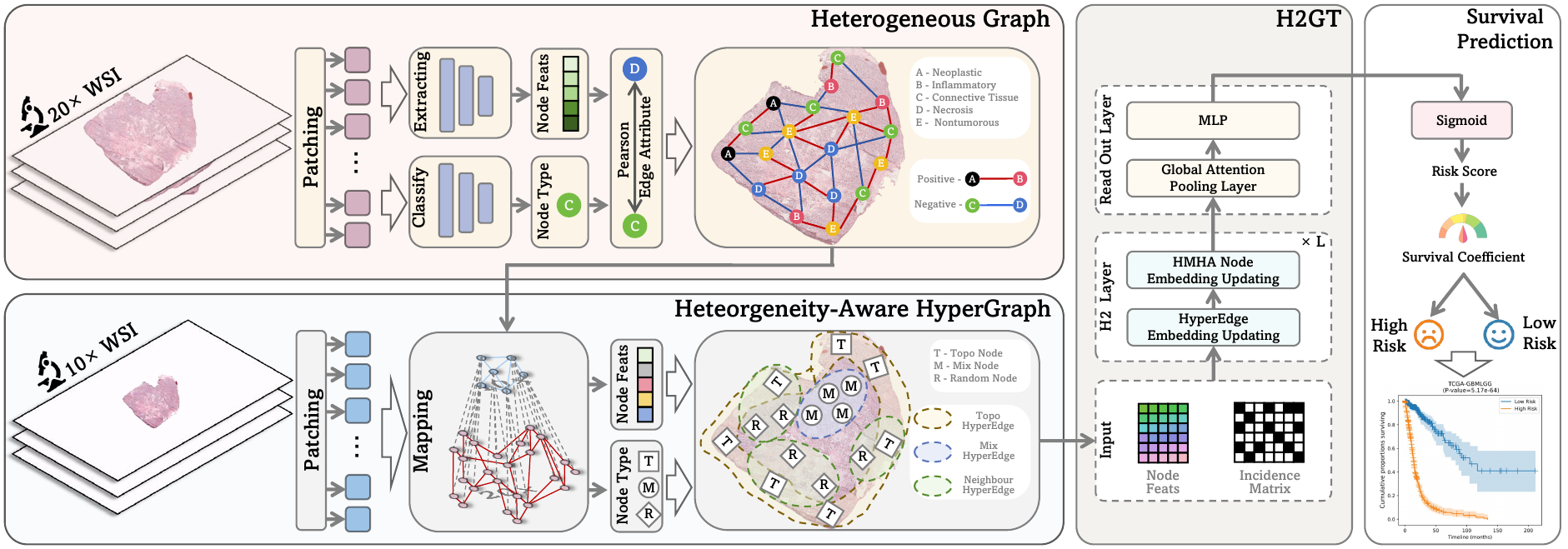}
    \caption{The overflow of our proposed multi-scale heterogeneity-aware hypergraph representation framework, which includes heterogeneous graph construction, heteorgeneity-aware hypergraph construction, and \textbf{H}eterogeneous \textbf{H}yper\textbf{G}raph \textbf{T}ransformer (H2GT).}
    \label{fig2}
\end{figure*}

The vast dimensions of WSIs, commonly sized at 40,000 × 40,000, render traditional methods used in natural image analysis ineffective. Consequently, Multiple instance learning (MIL) has become prevalent in most studies, where numerous patches are extracted as independent instances from WSIs and subsequently aggregated using a pooler for analysis. 
These methods do not require complex annotations and have made significant strides in cancer diagnosing and staging, as cancer diagnosing and staging only necessitate instance-level patch-based features~\cite{liDualstreamMultipleInstance2021a, gamper2021multiple, gamperPanNukeOpenPanCancer2019, qu2022dgmil, qu2022bi}. However, in contrast to them, survival prediction requires considering the interrelation between instances of tumors and surrounding tissues, as well as WSI's global features~\cite{shao2023characterizing,qu2024rise,qu2023rethinking}.
Yet, MIL-based methods encounter difficulties in effectively capturing the contextual information of WSIs. For example, though MIL-based methods would be able to learn instance-level features that discriminate image patches of stroma and tumor cells, it is unable to distinguish whether those cells have tumor-stromal infiltration. To address this limitation, researchers have proposed employing graph neural networks (GNNs) to characterize WSIs and capture contextual relationships among instances~\cite{nakhli2023sparse, guan2022node,di2022big}. Nevertheless, each node on the homogeneous graphs is considered ``equivalent", which is evidently incorrect when dealing with pathological slides. Pathological slides encompass a multitude of distinct tissue blocks, each exhibiting varying interrelations. As an example, Shao et al.~\cite{shao2023characterizing} highlighted that the frequent interaction between tumor tissues and immune tissues in pathological slides is advantageous for patient prognosis. Contrarily, Bhowmick et al.~\cite{bhowmick2005tumor} argue that there exist interactions between stromal cells and tumor cells, wherein stromal cells play a pivotal role in promoting the survival and proliferation of tumor cells. This is achieved through the secretion of extracellular matrix proteins, growth factors, and other substances by stromal cells. Moreover, homogeneous graph construction fails to simulate the diverse interactions among multiple tissue blocks within the global context of pathological slides.

To eliminate this problem, we propose a novel framework for survival prediction based on WSIs, which can leverage the interrelationships among different cell nucleus types and interactions across various tissues. The proposed framework consists of three key components: i) we constructed a heterogeneous graph neural network, assigning different categories to each patch and different types to edges connecting patches of different categories, to model the different kinds of interactions between distinct cell clusters; ii) to learn connections within and between tissue regions at larger scales, we build a heterogeneity-aware hypergraph on top of the heterogeneous graph. As shown in Figure \ref{fig1}, the heterogeneity-aware hypergraph allows for the modeling of more complex relationships and connectivity patterns, and iii) to characterize the interactions between different types of tissues and the embedding of the whole WSI, we propose a \textbf{H}eterogeneous \textbf{H}yper\textbf{G}raph \textbf{T}ransformer (H2GT), which provides different learnable projection matrices for different node types to increase the specificity and accuracy of information propagation. Additionally, extensive experiments were conducted on three datasets from The Cancer Genome Atlas (TCGA), and it was experimentally demonstrated that our framework outperforms current state-of-the-art (SOTA) methods.

\section{methodology}
\label{methods}
The overall framework of our proposed method is illustrated in Figure \ref{fig2}, which aims to estimate the survival coefficient $S_n(t|D_n)$ of patient $n$ based on clinical data $D_n$. The patient's clinical data is represented as $D_i=\left \{ P_i, c_i, t_i \right \} $, where $P_i$ stands for the patient's WSIs, $c_i \in \{0,1\}$ denotes whether the outcome event is right-censored ($c = 1$, patients were still alive at the end of the observation period) or not ($c = 0$), and $t \in \mathbb{R} ^+$ signifies the overall survival time. Subsequently, a hazard function is estimated: $h_n(t|D_n) = h_n(T = t|T \geq t, D_n) \in [0, 1]$. This function represents the likelihood of a death event occurring around time point $t$. Survival prediction does not involve directly estimating the overall survival time but instead utilizes the cumulative distribution function to output an ordered risk value to get the survival coefficient:$S_n(t|D_n)=\prod_{u=1}^{t}(1-h_n(u|D_n)) $, detailed information is available in the \textbf{Supplementary Materials}.

\subsection{Pairwise Heterogeneous Graph Construction}
\label{hetgraph construction}
A heterogeneous graph $\mathcal{G}'=\left \{ \mathcal{V}', \mathcal{E}', \mathcal{A}',\mathcal{R}'\right \} $ is defined as a graph with multiple node types $\mathcal{A}'$ or edge types $\mathcal{R}'$.  An edge $e'=(s,r',t)\in \mathcal{E}'$ links the source node $s$ and the target node $t$. Each node $v'$ has a $d$-dimensional node feature $x' \in X'$, where $X'$ is the embedding space of node features. All operations related to the pairwise heterogeneous graph were performed at $20\times$ magnification.

\textbf{Node Feature Extraction.} \,For a WSI image, we implement automatic tissue segmentation in two steps: 1) separating tissues and background using the OTSU thresholding algorithm, and then 2) using slide window strategy to generate multiple non-overlapping $256\times256$ pixels instance-level patches. These patches serve as the nodes to construct the heterogeneous graph. Subsequently, the CTransPath feature extractor~\cite{wangTransformerbasedUnsupervisedContrastive2022} is employed to extract features from each patch, resulting in $768$-dim embeddings.

\textbf{Heterogeneous Graph Construction.} \,After completing node feature extraction, the subsequent steps involve defining node types, edges, and edge attributes. We use the HoverNet~\cite{grahamHoverNetSimultaneousSegmentation2019a} pre-trained on the PanNuke dataset~\cite{gamperPanNukeOpenPanCancer2019} to categorize each patch into predefined classes as the node type. All patches are categorized into six distinct classes: \textit{Neoplastic}, \textit{Inflammatory}, \textit{Connective Tissue}, \textit{Necrosis}, \textit{Nontumorous}, and \textit{No label}. Subsequently, we applied the k-nearest neighbor algorithm to identify the $k'$ closest nodes for each node, based on their 2D spatial positions. Then, we calculated the Pearson correlation coefficient between the features of connected nodes for each edge. The edge attributes were labeled as ``positive" if the coefficient was positive and ``negative" otherwise. Data augmentation is employed to prevent noise during graph construction (\textit{i.e.}, random node features dropping and edges features dropping). As a result, we obtain a heterogeneous graph $\mathcal{G'}$ with heterogeneity due to the different node types and edge attributes.

\begin{figure}[tb]
    \centering
    \includegraphics[width=1\columnwidth]{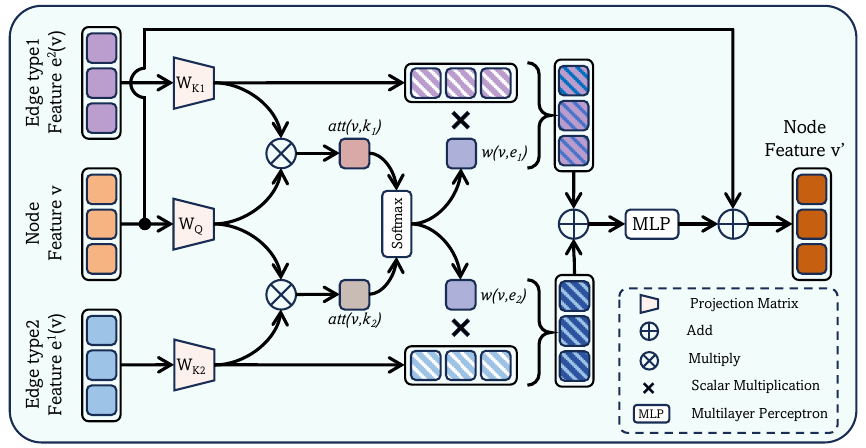}
    \caption{An example of a Heterogeneous Multi-Head Attention (HMHA) node embedding update module, where node $v$ is associated with $e^1(v)$ and $e^2(v)$, those hyperedges belong to two distinct categories.}
    \label{fig3}
\end{figure}

\subsection{Heterogeneous HyperGraph Construction}
\label{H2 construction}
Correspondingly, we use $\mathcal{G}=\{\mathcal{V}, \mathcal{E}, \mathcal{A}, \mathcal{R}, \mathcal{W}\}$ to denote heterogeneous hypergraph. The set of nodes and hyperedges are represented by $\mathcal{V} = \left \{  v_1,...,v_N\right \} $ and $\mathcal{E} = \left \{  e_1,...,e_M\right \} $, respectively, where $N$ and $ M $ are the number of nodes and hyperedges. $\mathcal{A}$ and $\mathcal{R}$ represent the sets of different types of nodes and hyperedges. Immediately, a positive diagonal matrix $\mathcal{W} \in \mathbb{R}^{N \times N}$ is utilized to represent the importance of hyperedges. For simplicity, we set this matrix as a diagonal all-one matrix. Also, we use an incidence matrix  $\textit{H} \in \mathbb {R} ^{N\times M}$ to indicate the relationships between nodes and hyperedges:
\begin{align}
h(v,e) & = \left\{
\begin{array}{cl}
1, &  if \ v \in e, \\
0,  &   otherwise,
\end{array} \right.
\end{align}
where $v$ and $e$ denote node and hyperedge separately.

We construct the hypergraph at $10\times$ magnification, thus one patch at $10\times$ magnification($Patch_{10\times}$) contains four patches at $20\times$ magnification($Patch_{20\times}$). Three node types, namely \textbf{Mix Node}, \textbf{Boundary Node}, and \textbf{Random Node}, are established within the heterogeneous hypergraph. Prior studies have demonstrated that tumor-stroma invasion profoundly impacts cancer progression, metastasis, and prognosis~\cite{hanahan2011hallmarks}~\cite{kather2020development}. Therefore, we set up Mix Node to explore the impact of tumor-stroma region on patient prognosis. A Mix Node contains \textit{Neoplastic} and \textit{Connective Tissue} typed $Patch_{20\times}$. Boundary Nodes are nodes sampled according to the topological structure of pathological images. Following that, an additional 500 $Patch_{10\times}$ are randomly sampled as Random Nodes. The feature for each node is obtained by averaging the features of its four containing $Patch_{20\times}$ after heterogeneous graph learning. For hyperedges, we also set three types. \textbf{Mix Hyperedge} contains all Mix Nodes. \textbf{Boundary Hyperedges} connect tissue boundary nodes. The coverage of Boundary Node and Boundary Hyperedges is presented in the \textbf{Supplementary Materials}. The k-nearest neighbor algorithm based on feature distance metrics is employed to construct the \textbf{Neighbour Hyperedges} for each node.

\subsection{Heterogeneous HyperGraph Transformer}
\label{H2GT}
Heterogeneous HyperGraph Transformer(H2GT) predicts the survival coefficient of each patient by analyzing the graph we have constructed; it mainly consists of two parts: i) the \textbf{H}eterogeneity-aware \textbf{H}ypergraph learning layer (H2 Layer), which comprises the hyperedge updating module and Heterogeneous Multi-head Attention (HMHA) node embedding updating module, and ii) the Read Out Layer.
\subsubsection{H2 Layer}

\textbf{Initial Node Embedding On HyperGraph.} To establish node embeddings for the hypergraph, we begin by inputting the heterogeneous graph into the network to initially discern the interconnections among diverse biological entities on a low scale. Here, we opt for R-GCN~\cite{schlichtkrullModelingRelationalData2018}, a widely employed method for heterogeneous graph learning. 
Then, we can obtain the features of hypergraph nodes by calculating the average of the features of the corresponding four $Patch_{20\times}$.

\textbf{Hyperedge Embedding Updating.} The embeddings of the nodes within each hyperedge are aggregated to derive embedding representations for the hyperedges. The hyperedges embeddings on the $l$-th layer are denoted as $E^l \in \mathbb{R}^{d \times M}$, which can obtained as follows:
\begin{align}
E^l(e^j(v_i)) & = \frac{X^l\cdot H(e^j(v_i))}{\delta(e^j(v_i)) },
\end{align}
$H(e^j(v_i)) \in \mathbb{R}^{N}$ is a column of $H$ and $\delta(e^j(v_i))$ is the degree of edge $e^j(v_i)$, it is used here as a normalization factor and is numerically equal to the sum of $H(e^j(v_i))$.

\begin{table*}[htb]
\caption{C-index($\uparrow$) performance(mean and standard deviation) comparisons of H2GT against prior state-of-the-art approaches on three datasets in the TCGA. The best and second best results are highlighted in \textcolor{red}{red} and \textcolor{blue}{blue}, respectively.}
\centering
\begin{tabular}{llclclclc} 
\hline
\multicolumn{1}{c}{\multirow{2}{*}{Model}} &  & \multicolumn{7}{c}{Datasets}                                                     \\ 
\cline{3-9}
\multicolumn{1}{c}{}                                  &  & BRCA                                &  & GBMLGG                           &  & BLCA                             &  & Overall     \\ 
\hline
AttnMIL~\cite{ilseAttentionbasedDeepMultiple2018}     &  & 0.5789 $\pm$ 0.0603                 &  & 0.7492 $\pm$ 0.0301              &  & 0.5778 $\pm$ 0.0533              &  & 0.6353      \\
DSMIL~\cite{liDualstreamMultipleInstance2021a}        &  & 0.5276 $\pm$ 0.1087                 &  & 0.7390 $\pm$ 0.0267              &  & 0.5629 $\pm$ 0.0685              &  & 0.6098      \\
CLAM-SB~\cite{lu2021data}                             &  & 0.5590 $\pm$ 0.0654                 &  & 0.7377 $\pm$ 0.0357              &  & 0.5856 $\pm$ 0.0284              &  & 0.6274      \\
CLAM-MB~\cite{lu2021data}                             &  & 0.5490 $\pm$ 0.0391                 &  & \color{red}{0.7505 $\pm$ 0.0356} &  & 0.5660 $\pm$ 0.0474              &  & 0.6218      \\ 
\hline
PatchGCN~\cite{chenWholeSlideImages2021}              &  & 0.5733 $\pm$ 0.1090                 &  & 0.7490 $\pm$ 0.0482              &  & \color{blue}{0.5993 $\pm$ 0.0495}&  & 0.6405      \\
GTNMIL~\cite{zhengGraphTransformerWholeSlide2022}     &  & 0.5762 $\pm$ 0.0728                 &  & 0.7432 $\pm$ 0.0362              &  & 0.5874 $\pm$ 0.0312              &  & 0.6356      \\ 
\hline
HEAT~\cite{chanHistopathologyWholeSlide2023}          &  & \color{blue}{0.6166 $\pm$ 0.0545}   &  & 0.7444 $\pm$ 0.0362              &  & 0.5871 $\pm$ 0.0498              &  & \color{blue}{0.6494}      \\
\rowcolor[rgb]{0.992,0.957,0.945} H2GT(Ours)   &  & \color{red}{0.6230 $\pm$ 0.0383}    &  & \color{blue}{0.7505 $\pm$ 0.0439}&  & \color{red}{0.6089 $\pm$ 0.0360} &  & \color{red}{0.6608}      \\
\hline
\end{tabular}
\label{对比1}
\end{table*}

\textbf{HMHA Node Embedding Updating.} \,Previous works calculate the similarity among hyperedges as weights for updating node embeddings~\cite{wangDynamicWeightedHypergraph2023}. However, they do not consider the diverse contributions of different hyperedges; that is, they ignore the heterogeneity of the hypergraph. To cope with that, an HMAT module is devised to learn the relative importance of various hyperedges concerning a node. We update the embeddings for node $v_i$ using all hyperedges related to $v_i$. The set of hyperedges associated with node $v_i$ is denoted as: $\mathcal{E}(v_i)=\left \{ e^1(v_i),...,e^{M_i}(v_i) \right \}$. $M_i$ is the number of hyperedges associated with node $v_i$.

Figure \ref{fig3} illustrates the scenario in which node $v$ is solely associated with $e^1(v)$ and $e^2(v)$, which belong to two distinct categories. We begin by feeding the node and hyperedge features into learnable, type-specific projection matrices to obtain query vector $Q^h(v_i)$ and key vector $K^h(e^j(v_i))$:
\begin{align}
Q^h(v_i) & = Q\text{-}Proj^h_{\phi (v_i)}\cdot X^l(v_i), \\
K^h(e^j(v_i)) & = K\text{-}Proj^h_{\psi  (e^j(v_i))}\cdot E^l(e^j(v_i)),
\end{align}
where $Q\text{-}Proj^h_{\phi (v_i)} \in \mathbb{R}^{\frac{d}{K} \times d}$ represents the projection matrix on the $h$-th attention head, $Q^h(v_i)$ and $K^h(e^j(v_i))$ are $\mathbb{R}^{\frac{d}{K}}$. $\phi (v_i)$ and $\psi  (e^j(v_i))$ are functions that respectively map nodes and hyperedges to their corresponding node types and hyperedge types, which means $\phi (v_i) \in \mathcal{A}$ and $\psi  (e^j(v_i)) \in \mathcal{R}$.
\begin{align}
\label{eq:att}
att^{h}(v_i,e^j(v_i)) & = \frac{Q^h(v_i)\cdot K^h(e^j(v_i)) \cdot \lambda_{\psi(e^j(v_i))} }{\sqrt{d} }.
\end{align}
Eq.~(\ref{eq:att}) derives the attention value of the $h$-th head between $v_i$ and $e^j(v_i)$. $\lambda $ is a type-specific scaling factor.

Subsequently, the $softmax$ function is applied to acquire the ultimate attention scores. These attention scores represent the relative importance of the incoming nodes for all related hyperedges:
\begin{align}
\label{eq:atten}
w^h(v_i,e) & = \mathop{softmax}\limits_{\forall{e\in \mathcal{E}(v_i) }}\left (  att^h(v_i,e)\right ),
\end{align}
where $w^h(v_i,e) \in \mathbb{R}^{M_i}$ is the final weight vector for each hyperedge on $h$-th attention head, and $\mathcal{E}(v_i)$ is the set of all related hyperedges. The aggregation of $h$-th attention head for node $v_i$ is formulated as:
\begin{align}
z^h_i & = \sum_{j \in \left [  1, M_i\right ] }^{}\left ( w^h(v_i,e^j(v_i)\times K^h(e^j(v_i)) \right ).
\end{align}

At last, after combining all $K$ heads, the updated node embedding of $v_i'$ is derived by passing it through a Multilayer Perceptron (MLP) featuring a weighted skip connection.

\subsubsection{Read Out Layer} Following~\cite{ilseAttentionbasedDeepMultiple2018}, we employ a global attention-based pooling layer $\mathcal{F}_{Atten}(X^L)$, to dynamically calculate a weighted sum of node features in the graph, $L$ denotes the total number of layers in the H2 Layer. The pooling process converts the node feature matrix $X^L\in\mathbb{R}^{N \times d}$ from the final layer into the WSI-level embedding representation $h^L\in\mathbb{R}^d$. Ultimately, $h^L$ undergoes an MLP to derive the final risk scores. The risk scores are then subjected to supervision by applying the NLL-loss~\cite{zadehBiasCrossEntropyBasedTraining2021a}, and additional details about the survival loss function can be found in the \textbf{Supplementary Materials}.

\section{EXPERIMENTS}
\subsection{Implementation Details}
The Adam optimizer is employed with a learning rate of 1e-4, a weight decay of 1e-5, and the model is trained for 20 epochs. Furthermore, we apply data augmentation by randomly dropping node and edge features with a dropout rate of 0.3. Hyperparameters $K$ and $L$ are set to 4 and 2, respectively. For a more equitable evaluation of different methods, identical loss functions and feature extractors are used across all methods. Since the graph sizes vary greatly, we set the batch size to 1, with 32 gradient accumulation steps. All other details will be available in our code. In order to maximize the utilization of the data and enhance the robustness of the model, 5-fold cross-validation was employed.

\subsection{Datasets and Evaluation Metrics}
\textbf{Datasets.} For a more compelling experimental setup, we leverage three datasets sourced from TCGA: Bladder Urothelial Carcinoma (BLCA) (n = 385), Breast Invasive Carcinoma (BRCA) (n = 793), Glioblastoma \& Lower Grade Glioma (GBMLGG) (n = 795). When a patient possesses multiple WSIs, only the largest WSI is selected for analysis. As a result, there are 1973 WSIs used for training and testing, with an average of 12,692 patches per WSI (totaling approximately 25 million patches, 2TB).

\noindent \textbf{Evaluation Metric.} The concordance index (C-index) is a statistical metric commonly used to evaluate the predictive accuracy of survival analysis models. It measures the ability of a model to correctly rank pairs of samples based on their predicted probabilities of experiencing an event (\textit{e.g.}, death) over a given time period. We start by sorting all the samples in ascending order based on their predicted survival probabilities,
then the C-index can be formulated as follows:
\begin{align}
C\text{-} index  = \frac{1}{N(N-1)}\sum_{i  = 1}^{N}  \sum_{j = 1}^{N}\mathbb{I}(T_i< T_j)(1-c_j),
\end{align}
where $N$ is the number of patients and $\mathbb{I}(\cdot)$ is the indicator function.

\subsection{Comparisons with other methods}
To show the superior performance of our proposed method, we select a series of SOTA methods for comparison. They include MIL-based methods: \textbf{AttnMIL}~\cite{ilseAttentionbasedDeepMultiple2018}, \textbf{DSMIL}~\cite{liDualstreamMultipleInstance2021a}, \textbf{CLAM-MB}, and \textbf{CLAM-SB}~\cite{lu2021data}; homogeneous graph-based methods: \textbf{PatchGCN}~\cite{chenWholeSlideImages2021}, \textbf{GTNMIL}~\cite{zhengGraphTransformerWholeSlide2022}; and heterogeneous graph-based method: \textbf{HEAT}~\cite{chanHistopathologyWholeSlide2023}.
\begin{table}
\caption{Comparison of H2 Layer with other hypergraph learning networks.}
\resizebox{\columnwidth}{!}{
\centering
\setlength{\extrarowheight}{0pt}
\addtolength{\extrarowheight}{\aboverulesep}
\addtolength{\extrarowheight}{\belowrulesep}
\setlength{\aboverulesep}{0pt}
\setlength{\belowrulesep}{0pt}
\begin{tabular}{llccc} 
\toprule
\multirow{2}{*}{Model} & \multirow{2}{*}{} & \multicolumn{3}{c}{Datasets}                   \\ 
\cline{3-5}
                                                            &                   & BRCA                              & GBMLGG                         & BLCA                 \\ 
\hline
MLP                                                         &                   & 0.5613 $\pm$ 0.0338               & 0.7358 $\pm$ 0.0283            & 0.5657 $\pm$ 0.0465  \\
HGNN~\cite{hgnn}                                            &                   & 0.5757 $\pm$ 0.0974               & \textbf{0.7544 $\pm$ 0.0273}   & 0.5962 $\pm$ 0.0397  \\
HyperGCN~\cite{hypergcn}                                    &                   & 0.5649 $\pm$ 0.0572               & 0.7465 $\pm$ 0.0304            & 0.4534 $\pm$ 0.0638  \\
HyperGAT~\cite{hypergat}                                    &                   & 0.6206 $\pm$ 0.0623               & 0.7443 $\pm$ 0.0230            & 0.5610 $\pm$ 0.0576  \\
\rowcolor[rgb]{0.992,0.957,0.945} H2 Layer(Ours)                &                   & \textbf{0.6230 $\pm$ 0.0383}      & 0.7505 $\pm$ 0.0439   & \textbf{0.6089 $\pm$ 0.0360}  \\
\bottomrule
\end{tabular}
}

\label{对比2}
\end{table}

\begin{table}
\caption{Ablation study assessing the impact of three modules in H2GT.}
\resizebox{\columnwidth}{!}{
\centering
\setlength{\extrarowheight}{0pt}
\addtolength{\extrarowheight}{\aboverulesep}
\addtolength{\extrarowheight}{\belowrulesep}
\setlength{\aboverulesep}{0pt}
\setlength{\belowrulesep}{0pt}
\begin{tabular}{ccclccc} 
\hline
\multirow{2}{*}{TSP} & \multirow{2}{*}{Mix} & \multirow{2}{*}{Bdy} & \multirow{2}{*}{} & \multicolumn{3}{c}{Datasets}                         \\ 
\cline{5-7}
                                                &             &             &   & BRCA                              & GBMLGG                            & BLCA             \\ 
\hline
-                                               & \checkmark  & \checkmark  &   & 0.5910 $\pm$ 0.0604               & 0.7458 $\pm$ 0.0429               & 0.5870 $\pm$ 0.0350  \\
\checkmark                                      & -           & \checkmark  &   & 0.6056 $\pm$ 0.0394               & 0.7383 $\pm$ 0.0420               & 0.5609 $\pm$ 0.0631  \\
\checkmark                                      & \checkmark  & -           &   & 0.6108 $\pm$ 0.0467               & 0.7376 $\pm$ 0.0394               & 0.5844 $\pm$ 0.0701  \\
\rowcolor[rgb]{0.992,0.957,0.945}\checkmark         & \checkmark  & \checkmark  &   & \textbf{0.6230 $\pm$ 0.0383}      & \textbf{0.7505 $\pm$ 0.0439}      & \textbf{0.6089 $\pm$ 0.0360}  \\
\hline
\end{tabular}
}

\label{消融}
\end{table}
Table \ref{对比1} indicates that H2GT achieves an average C-index of 66.08\%, surpassing all other advanced methods. Specifically, H2GT outperforms all prior approaches on BRCA and BLCA and ties for first place with CLAM-MB on GBMLGG. Compared to the best-performing MIL-based, homogeneous graph-based, and heterogeneous graph-based methods, H2GT demonstrates superior performance, improving average C-index by 4.01\%, 3.17\%, and 1.76\%, respectively. This shows that i) graph structures may be more capable of capturing global information in WSIs compared to MIL-based methods, ii) heterogeneous graph modeling approaches can better represent interactions between patches in WSIs than homogeneous graph-based methods, and more importantly, iii) our heterogeneity-aware hypergraph construction method works best, indicating its ability to further uncover interrelationships between different biological entities globally in WSIs.

To further showcase the superior performance of our heterogeneity-aware hypergraph learning module, the H2 Layer is replaced with four other baseline models for hypergraph learning: 1) \textbf{MLP}: ignores hypergraph structure and only uses node features as input, 2) \textbf{HGNN}~\cite{hgnn}: Hypergraph Neural Networks, 3) \textbf{HyperGCN}~\cite{hypergcn}: graph convolutional-based hypergraph learning model, 4) \textbf{HyperGAT}~\cite{hypergat}: graph attention-based hypergraph learning model.

In Table \ref{对比2}, we evaluate the performance of four hypergraph learning modules in H2GT on three datasets. The H2 Layer outperforms MLP, HGNN, HyperGCN and HyperGAT by 6.43\%, 2.91\%, 12.32\%, and 2.93\% in terms of average C-index.
This validates that considering node heterogeneity and hyperedge heterogeneity in hypergraph learning can more effectively represent the global information of WSIs.
\begin{figure}[tb]
    \centering
    \includegraphics[width=1\columnwidth]{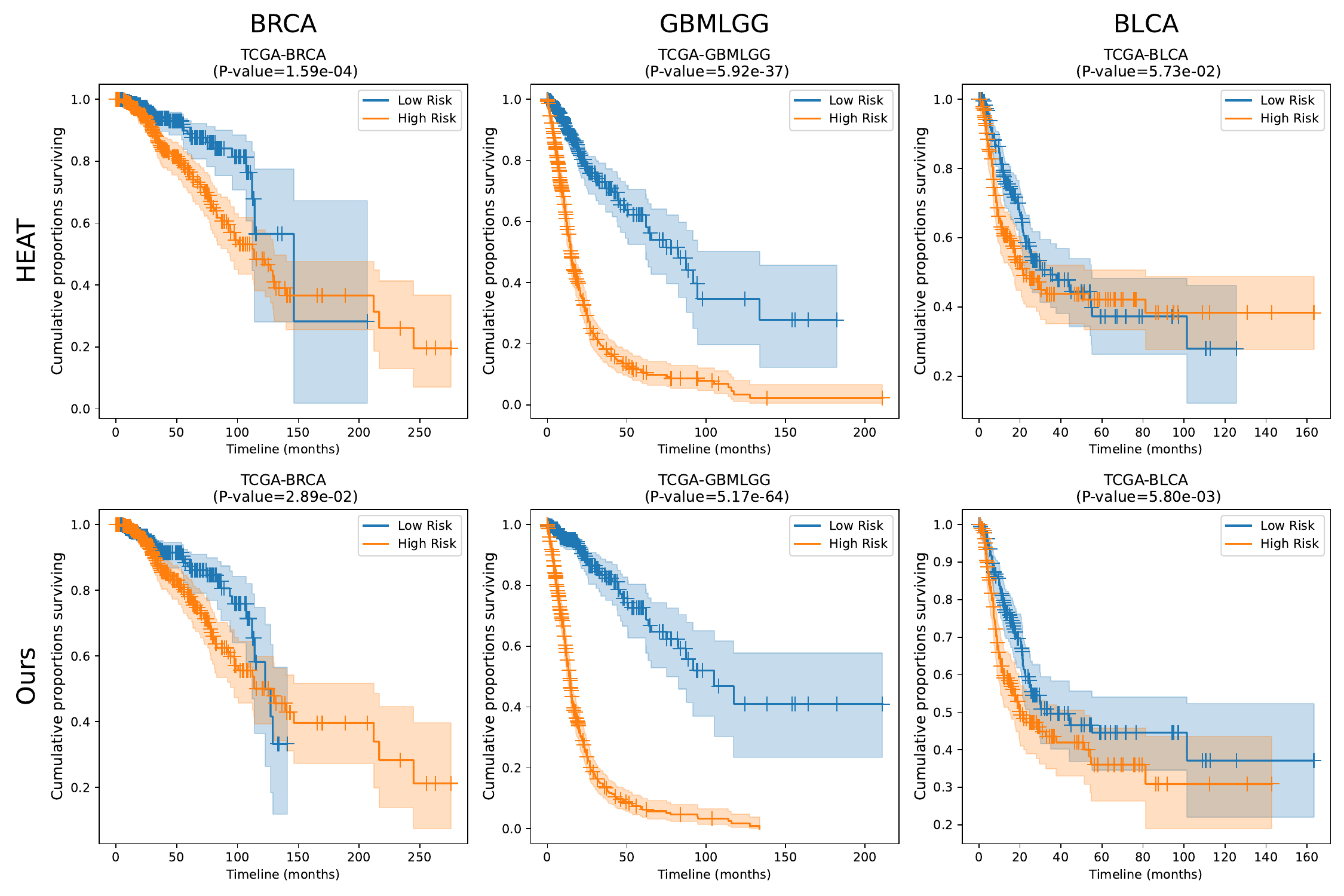}
    \caption{Kaplan-Meier Analysis on three datasets, where patient stratifications of low risk (\textcolor[RGB]{31,119,180}{blue}) and high risk (\textcolor[RGB]{255,127,14}{orange}) are presented. Shaded areas refer to the confidence intervals. P-value $<$ 0.05 means the significant statistical difference in two groups, and the lower P-value is better.}
    \label{fig4}
\end{figure}

\subsection{Ablation Studies}
An ablation study was conducted to assess the influence of type-specific projection (TSP), Mix node (Mix), and Boundary node (Bdy) on model performance by examining changes in H2GT upon the removal of each module. Table \ref{消融} shows the experimental results. Notably, removing TSP means the three node types share the same projection matrix when updating node embeddings. Besides, when Mix or Bdy is removed, they are replaced with an equivalent number of Random Nodes to ensure fairness. Experiments demonstrate that exploring heterogeneity in hypergraphs is necessary and useful for survival prediction, as removing any component has a negative impact on the performance.
\subsection{Patient Stratification}
Patient stratification plays a crucial role in tailoring treatment plans for patients, benefiting their clinical outcomes significantly. Therefore, we plotted the Kaplan-Meier survival curve to demonstrate the effectiveness of the H2GT method for patient stratification. Figure \ref{fig4} presents the results of H2GT and HEAT on three datasets alongside their P-values. The Logrank test shows that our model is statistically significant (p$<$0.05) and outperforms most models across all datasets. The Kaplan-Meier survival curves of other methods can be found in the \textbf{Supplementary Materials}.

\section{CONCLUSION}
In this paper, we propose a heterogeneity-aware hypergraph representation framework for patient survival prediction. This framework can model WSIs at both high and low scales, analyzing interactions between different types of biological entities. Experiments on three datasets demonstrate that compared with other SOTA methods, our proposed framework can better uncover global information in WSIs and achieve superior performance on survival prediction and patient stratification. Overall, this paper proposes a novel representation learning framework for WSIs and demonstrates its effectiveness in clinically relevant tasks such as survival prediction and patient stratification.

\section*{Acknowledgment}

This project was funded by the National Natural Science Foundation of China 82090052.

\bibliographystyle{IEEEbib}
\bibliography{ICME2024}

\newpage
\section{Appendix}
\subsection{Survival Analysis}
\textbf{Preliminaries} \, Survival prediction is an ordinal regression task that models time-to-event distributions, where the outcome of the event is not always observed (\textit{e.g.} right censored). In observational studies that examine overall survival in cancer patients, a censored event would result from the last known patient follow-up, while an uncensored event would be observed as patient death.

Following our notation in \cref{methods}, let $P$ represent patient's pathological image, $t \in \mathbb{R} ^+$ signifies the overall survival time, $c_i \in \{0,1\}$ denotes whether the outcome event is right-censored $(c = 1)$ or not $(c = 0)$. In addition, let $T$ be a continuous random variable for overall survival time, the survival function $S(t|D)$ measures the probability of patient $n$ surviving longer than a discrete time point $t$, and the hazard function $h(t|D) = h(T = t|T \geq t, D)$ measure the probability of patient death instantaneously at $t$, defined as:
\begin{align}
h(T = t)  = \lim_{\partial t \to 0}\frac{P(t\le T \le t+\partial t|T \ge t)}{\partial t}, 
\end{align}
which can be used to estimate $S_n(t|D_n)$ by integrating over of $h$. The most common method for estimating the hazard function is the Cox Proportional Hazards (CoxPH) model, in which $h$ is parameterized as an exponential linear function,
\begin{align}
h(t|D) & = h_0(t)e^{\theta D},
\end{align}
 where $h_0$ is the baseline hazard function and $\theta$ are model parameters that describe how the hazard varies with data $D$.  Using deep learning, $\theta$ is the last hidden layer in a neural network and can be optimized using Stochastic Gradient Descent with the Cox partial log-likelihood.
 
\textbf{Weak Supervision with Limited Batch Sizes} \, A second approach to survival prediction using deep learning is to consider discrete time intervals and model each interval using an independent output neuron. This formulation overcomes the need for large mini-batches and allows the model to be optimized using single observations during training. Specifically, given right-censored survival outcome data, we build a discrete-time survival model by partitioning the continuous time scale into non-overlapping bins: $[t_0,t_1), [t_1,t_2), [t_2,t_3), [t_3,t_4)$ based on the quartiles of survival time values of uncensored patients in each TCGA cohort. The discrete event time of each patient, indexed by $j$, with continuous event time $T_{j,cont}$ is then defined by:
\begin{align}
T_j & = r \ \mathrm{if} \ T_{j,cont} \in [t_r,t_{r+1}) \ \mathrm{for} \ r\in \left \{ 0,1,2,3 \right \}.
\end{align}
Given the discrete-time ground truth label of the $j_{th}$ patient as $Y_j$. For a given patient with bag-level feature $h^L_j$ , the last layer of the network uses the sigmoid activation and models the hazard function defined as:
\begin{align}
h(r|h_j^L)  = P(T_j  = r|T_j \ge r, h^L_j),
\end{align}
which relates to the survival function through:
\begin{align}
S(r|h^L_j) &= P(T_j > r|h^L_j)\\
                  &=\prod_{u=1}^{r}(1-h(u|h^L_j)).
\end{align}
\subsection{Loss Function}
During training, we update the model parameters using the log-likelihood function for a discrete survival model, taking into account each patient’s binary censorship status ($c_j = 1$ if the patient lived past the end of the follow-up period and $c_j=0$ for patients who passed away during the recorded event time $T_j$):
\begin{align}
L = -l & = -c_j\cdot (S(Y_j|h_j^L)) \nonumber\\
       &- (1-c_j) \cdot log(S(Y_j-1|h^L_j)) \\
       &-(1-c_j) \cdot log(S(Y_j|h^L_j)). \nonumber
\end{align}
During training, we additionally up-weight the contribution of uncensored patient cases using a weighted sum of $L$ and $L_{uncensored}$. In this article, in order to ensure a fair comparison of the performance of each method, we uniformly set $\beta$ to 0.4.
\begin{align}
L_{surv} & = (1-\beta ) \cdot L+ \beta \cdot L_{uncensored}.
\end{align}
The $L_{uncensored}$ is computed as:
\begin{align}
L_{uncensored}= &-(1-c_j) \cdot log(S(Y_j-1|h^L_j)) \nonumber  \\
               &-(1-c_j) \cdot log(h(Y_j|h^L_j)).
\end{align}
\begin{figure*}[!htp]
    \centering
    \includegraphics[width=2.0\columnwidth]{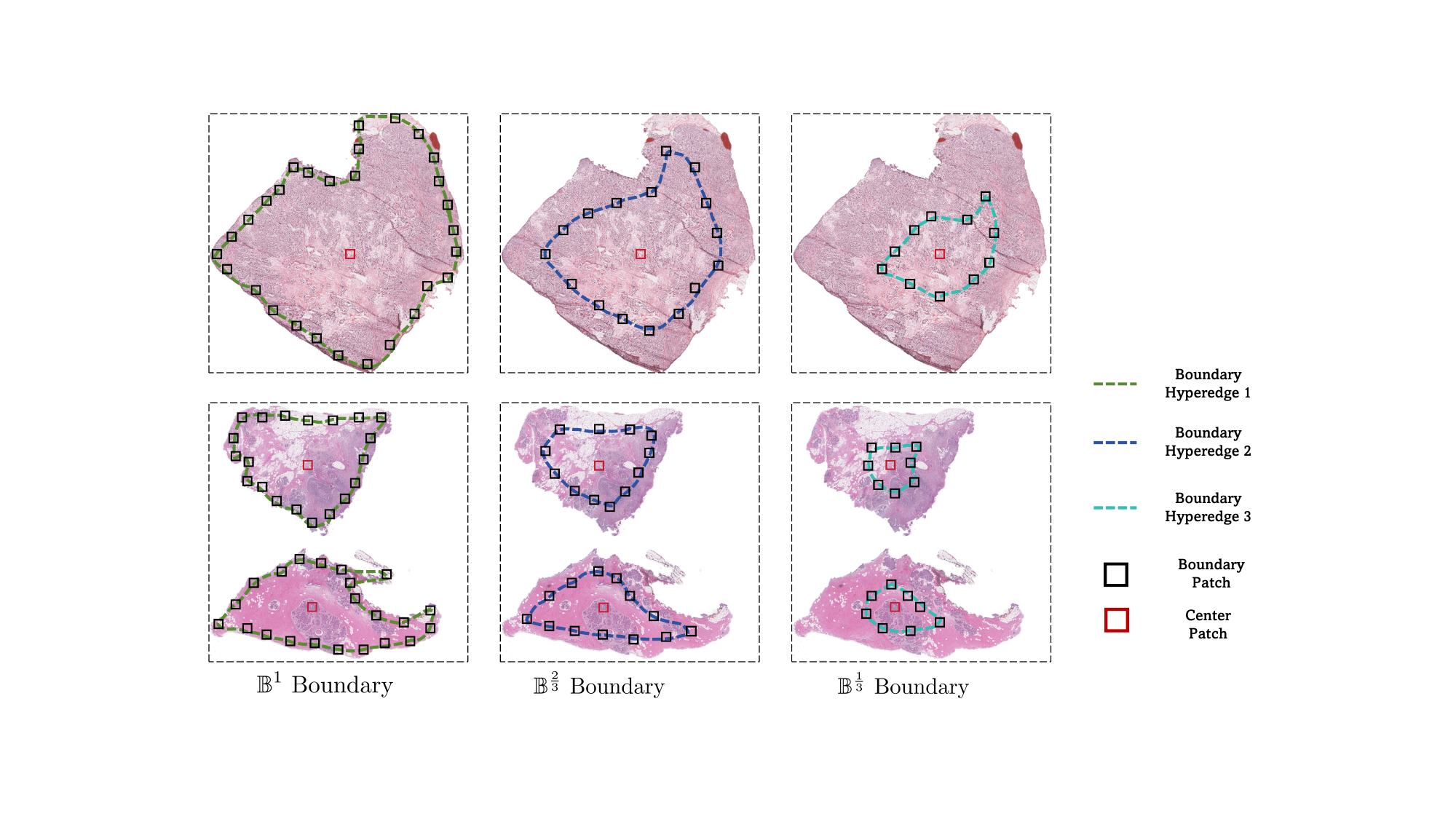}
    \caption{We constructed three Boundary Hyperedges for each WSI according to the distance of each node from the \textcolor[RGB]{192,0,0}{central point $\mathbb{C}$}, namely \textcolor[RGB]{88,142,49}{Hyperedge $\mathbb{B} ^1$}, \textcolor[RGB]{46,84,161}{Hyperedge $\mathbb{B} ^{\frac{2}{3}}$} and \textcolor[RGB]{100,208,199}{Hyperedge $\mathbb{B} ^{\frac{1}{3}}$}.}
    \label{fig5}
\end{figure*}

\subsection{Boundary Nodes and Hyperedges}
It is indicated by the research that the random selection of patches from WSI can disrupt the original topological structure, potentially affecting prognostic information embedded within the WSI. However, the manual selection of patches by pathologists would entail a significant consumption of time and money. Therefore, the decision was made to select Boundary Nodes and Boundary Hyperedges within the WSI to preserve the topological information of regions of interest. Additionally, experimental results have shown that selecting Boundary Nodes and Boundary Hyperedges contributes to prognostic analysis tasks.

As mentioned in \cref{hetgraph construction}, the OTSU algorithm is employed for background differentiation within the tissue section. Following this, we delineate the boundary of individual tissue blocks, denoted as boundary $\mathbb{B}^1$. Additionally, we derive the "concentric sub-boundaries" of each tissue block located at distances of $\frac{1}{3}$ and $\frac{2}{3}$ away from its central point $\mathbb{C}$, identified as Boundary $\mathbb{B}^{\frac{1}{3}}$ and Boundary $\mathbb{B}^{\frac{2}{3}}$, respectively. Figure \ref{fig5} shows the process of selecting Boundary nodes and Boundary Hyperedges.

\subsection{Patient Stratification}
\begin{figure*}[!htp]
    \centering
    \includegraphics[width=2.0\columnwidth]{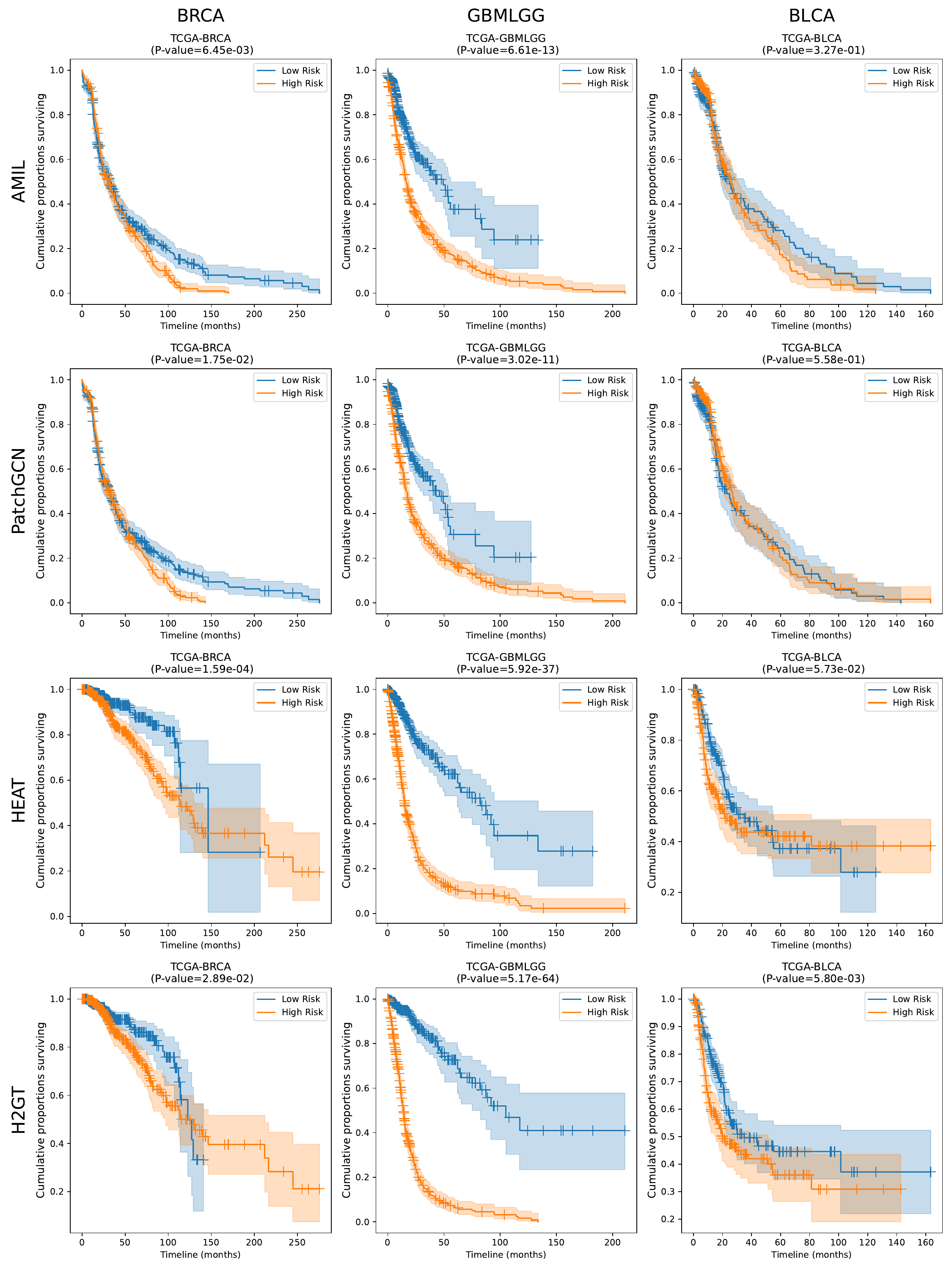}
    \caption{Kaplan-Meier Analysis on three datasets, where patient stratifications of low risk (\textcolor[RGB]{31,119,180}{blue}) and high risk (\textcolor[RGB]{255,127,14}{orange}) are presented. Shaded areas refer to the confidence intervals. P-value $<$ 0.05 means a significant statistical difference in two groups, and a lower P-value is better.}
    \label{fig6}
\end{figure*}

In Figure \ref{fig6}, we report the stratification results of H2GT(the best model), HEAT(the second best model), AttnMIL(the best MIL-based method), and PatchGCN(the best homogeneous graph-based method) on the three datasets and the corresponding P-values. The Logrank test shows that our model is statistically significant (p$<$0.05) and outperforms most models across all datasets (only inferior to HEAT on BRCA).

\end{document}